# Risk Prediction of Cardiovascular Disease for Diabetic Patients with Machine Learning and Deep Learning Techniques

Esha Chowdhury
*Dept. of CSE*
*Dhaka University of Engineering & Technology,*
Gazipur, Bangladesh
eshachowdhury.cse@gmail.com

*Abstract*— Accurate prediction of cardiovascular disease (CVD) risk is a crucial task for healthcare institutions. This work is important since diabetes is increasing and is strongly linked to heart disease. This study proposes an efficient CVD risk prediction model for diabetic patients using a combination of machine learning (ML) and hybrid deep learning (DL) approaches. The BRFSS dataset was preprocessed by removing duplicates, handling missing values, identifying categorical and numerical features, and applying Principal Component Analysis (PCA) for feature extraction. Several ML models, including DT, RF, KNN, SVM, AdaBoost, and XGBoost, were implemented, with XGBoost achieving the highest accuracy of 0.9050. Various DL models (ANN, DNN, RNN, CNN, LSTM, BiLSTM, and GRU) and hybrid models combining CNN with LSTM, BiLSTM, and GRU variants were also explored, with some achieving perfect recall (1.00), while the LSTM model achieved the highest accuracy of 0.9050. Our research highlights the effectiveness of ML and DL models in predicting cardiovascular disease risk among diabetic patients, thereby automating and enhancing clinical decision-making. High accuracy and F1 scores demonstrate these models' capability to improve personalized risk management and preventive strategies.

## I. Introduction

Increased prevalence of diabetes is leading to growth in cardiovascular disease in the whole world. This has added to the proactive risk management being a prerequisite health plan of all the people [1]. Due to the requirement to process a lot of complex patient data and profiles, healthcare practitioners cannot calculate or predict diabetes patients' heart disease risk in a timely manner [2]. Doctors in the past used to use traditional clinical scoring systems to ascertain the likelihood of a diabetic patient developing a heart attack or stroke. Such systems are not always absolutely correct and foresee the future events. More advanced procedures are thus desperately needed to ensure more accuracy and automation of cardiovascular risk prediction of this group of vulnerable population.

Techniques of Ensemble Machine Learning have changed healthcare [3]. We use a machine learning-based analytic platform to make predictions. We use machine learning (ML) and deep learning (DL) to acquire useful information. We analyze data related to patient health variables and biomarkers. We achieve this by using algorithms such as random forest, decision tree, and AdaBoost. This technology enables users to make better decisions and anticipates more robust decision-making systems. We enhance methods of precise cardiovascular risk assessment through model training and validation. We employ cutting-edge deep learning architectural guidelines to comply with clinical guidelines. We process measures of clinical data and standardize physiological data. We codify categorical health data and measure its accuracy. We produce a reliable dataset for cardiovascular risk modeling.

There is potential in deep learning in cardiology to enhance the process of cardiovascular risk assessment. These complex deep learning patterns can be studied using models applied to large, multidimensional datasets, which is typically challenging for traditional machine learning models [4]. This study will address the drawback of this, eliminating shortcomings of traditional methods with an advancement of a more intelligent and reconfigurable system that is capable of managing the dynamic nature of health data [5] and disease progression of diabetic patients. The study on DL is also aimed at increasing the clarity and cognition of cardiovascular risk forecasts and building confidence in the use of such sophistications in clinical diabetic patient care decision-making.

## II. Related Work

Several papers have examined the use of ML and DL models in predicting cardiovascular diseases, diabetic complications, and early intervention solutions. When the Linear Regression (LR) model was used to predict 10-year CVD risk, the highest results of 0.023 and 0.001 were achieved on the test set when utilizing the entire set of features [2]. Deep learning algorithms are built on a solid paradigm of prediction of the risk of coronary artery disease (CAD) and cardiovascular disease (CVD) risk classification [4]. Furthermore, meta-learning models, especially support vector machines (SVMs), have been developed to improve the prediction of CVD risk by identifying the complicated trends in the large datasets [6]. The effectiveness of such algorithms usually lies in the proper methods of selection of features, which play an important role in the accuracy and efficiency of predictions [7]. A Study indicates that the Deep Belief Network (DBN)-based model was found to have a better prediction accuracy of about 81.20% than traditional algorithms, including SVM and Artificial Neural Networks (ANN) models [8]. Basic risk factors had a greater predictive performance in nondiabetic women (AUC = 0.79) but had the same predictive performance in nondiabetic men (AUC = 0.69) [9]. Zarkogianni et al. [10] found that related risk prediction models based on hybrid wavelet neural networks and self-organizing maps had an AUC of 71.48.

## III. METHODS

This study evaluates the performance of various ML and DL Models for CVD risk prediction in diabetic patients. The methodology includes several key stages (see figure 1).

*A. Dataset Description*

The research is based on a large dataset from the Behavioral Risk Factor Surveillance System (BRFSS) to determine the risk of cardiovascular disease among diabetic patients. The data in the 2023 BRFSS dataset of the US Division of Population Health National Center of Chronic Disease Prevention and Health Promotion were used in the study. The table of attributes of the dataset is shown in TABLE II. The data was obtained on the Diabetes Health Indicators Dataset page in Kaggle. The dataset will also have lifestyle variables such as smoking and physical activity in addition to health variables, such as blood pressure and cholesterol levels. The most significant characteristics of the dataset are shown in TABLE III.

*B. Data Preprocessing*

The data was critically handled and normalized to prepare it to be used in learning ML and DL models. This entailed some of the key steps, such as eliminating duplication, missing values, categorizing columns as numerical and categorical, and dividing the data into training and test data, as illustrated in Fig. 1. The dataset that was preprocessed had 17 chosen features and 433,324 entries. It was further divided into 80:20 to have 346,242 samples of the training set and 86,662 of the testing set.

*C. Feature Extraction*

In an attempt to simplify the data, this research paper used Principal Component Analysis (PCA), which transforms variables into unrelated elements and preserves the most important patterns at the same time. The method is particularly beneficial in the analysis of medical data because the information about patients is considerably more comprehensible.

*D. Hyper Parameter Tuning*

In this study, we optimized our approach to accurately predict the likelihood of cardiovascular disease among diabetic patients by utilizing hyperparameter values and cross-validation techniques. We have tuned parameters in the machine learning models, including the regularization strength (C) and the type of kernel, to achieve optimal accuracy and generalization. The objective was to enhance the accuracy of prediction and not to overfit to the training data. Each of the models was tested with the help of the Grid Search CV and k-fold cross-validation with various combinations of parameters. The result was a hyperparameter optimization that was performed on all our models. The best models, after tuning in Colab on the BRFSS dataset, were the XGBoost and LSTM models. The higher accuracy scores reflect the significant increase in performance that our hyperparameter adjustments produced.

TABLE I
HYPERPARAMETER SETTING

| Epoch | Accuracy | Learning Rate | Optimizer |
|---|---|---|---|
| 8 | 95.00 | 0.01 | Adam |
| 8 | 95.00 | 0.001 | Adam |
| 8 | 95.00 | 0.0001 | Adam |

*E. Machine Learning Approach*

In this research, we investigated several machine learning models to forecast cardiovascular disease in diabetic patients. Decision Tree (DT) model is simple to comprehend, and it is applicable to both numerical and categorical data [11]. However, the Random Forest (RF) model is more capable of dealing with complex data and minimizing overfitting [11]. The K-Nearest Neighbors (KNN) algorithm ranks the data according to the proximity of data points, thereby acting well in identifying the complex patterns. AdaBoost is performed in stages; each time a new model is being trained, it corrects the errors made by the previous model. Finally, eXtreme Gradient Boosting (XGBoost), which integrates multiple classifiers, is especially good at boosting the performance of models and making them stronger.

*F. Deep Learning Approach*

The paper will analyze how deep learning models can be used to identify patients who are at risk of developing cardiovascular diseases when they have diabetes. CNN is particularly useful with grid-like data and is adept at finding patterns and features [12]. Artificial Neural Networks (ANN) are able to make forecasts, identify errors, and perform repetitive changes to their internal connections to enhance their performance. The Recurrent Neural Networks (RNN) are created with sequential dependencies and are especially applicable in predicting risk of disease. Long Short-Term Memory (LSTM) networks are better at dealing with long-term dependencies, and thus they are capable of storing important information that can be used to make accurate predictions. Deep Neural Networks (DNN) allow model learning of complex patterns of data that are on a multi-level, and they increase the accuracy of the model. Bidirectional LSTM (BiLSTM) is more reliable in recognizing the patterns as it processes the data in both directions. Finally, Gated Recurrent Units (GRU) are used to overcome the vanishing gradient issue and make models learn on long sequences.

*G. Hybrid Model*

To make better predictions, researchers have also looked into hybrid models that use more than one deep learning design. The study introduces numerous new hybrid deep learning algorithms for heart disease risk prediction. Each combination of CNN with LSTM, BiLSTM, and GRU, as well as LSTM with GRU and BiLSTM with GRU. It's beneficial that CNN can pick out features in space, and the recurrent networks make it more accurate, address the problem of disappearing gradients, and understand how things relate to each other over time. When looking for malware, for instance, a mix of LSTM and CNN has been used before [12]. Each hybrid model demonstrates its ability to enhance and streamline clinical decision-making. This makes it possible to create more personalized risk management and avoidance plans

## IV. RESULT AND ANALYSIS

### A. Confusion Matrix

A confusion matrix is a table that visualizes the performance of an algorithm by showing actual versus predicted classifications [6]. The confusion matrix and ROC curve results of the models are given below (see figures 2, 3, 4, 5, 6, 7, and 8):

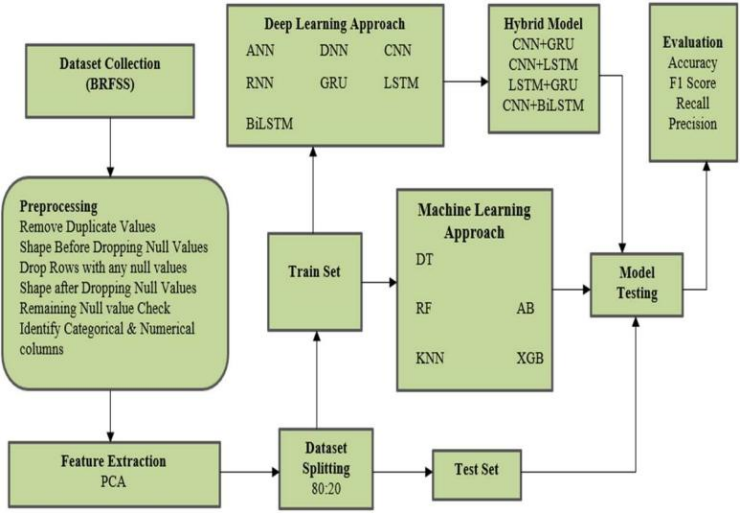

Fig. 1. Proposed Method

TABLE II

DETAILS OF THE 2023 BRFSS DATASET

| Attribute | Details |
|---|---|
| Title | 2023 BRFSS Survey Data |
| Author | U.S. Centers For Disease Control and Prevention |
| Features | 303 variables covering different aspects |
| Data Items | Responses from approximately 433,324 participants across the United States |
| Collection Method | Computer-Assisted Telephone Interview systems |
| Geographical Coverage | All 50 states of the United States, the District of Columbia, Guam, Puerto Rico, and US Virgin Islands |
| Categories of Features | Demographics, Chronic Health Conditions, Behavioral Features, Preventive Health Services, Physical and Mental Health Status, Disability Status and others |

TABLE III

KEY FEATURES OF THE DATASET

| Feature | Description |
|---|---|
| DIABTYPE | According to your doctor or other health professional, which type of diabetes do you have? |
| DIABETES | Diabetes is represented by three distinct values: 0 indicates no diabetes, 1 indicates pre-diabetes, and 2 indicates diabetes. |
| HighBP | Binary variable indicating high blood pressure. |
| HighChol | Binary variable indicating high cholesterol. |
| CHOLCHK | Indicates whether cholesterol levels have been checked within the past five years. |
| BMI | Body Mass Index; weight in kilograms divided by height in meters squared. |
| SMOKE | 0 indicates never smoked; 1 indicates having smoked at least 100 cigarettes in a lifetime. |
| CVDSTRK | Indicates whether the patient has ever had a stroke. |
| EXERANY | During the past month, other than your regular job, did you participate in any physical activities or exercises? |
| MEDCOST | During the past 12 months, was there a time when you needed to see a doctor but could not because of cost? |
| GENHLTH | General health status. Categories: Excellent, Very good, Good, Fair, and Poor. |
| MENTHLTH | The number of days in the past month when mental health was poor. |
| PHYSHLTH | The number of days in the past month when physical health was poor. |
| DIFFWALK | Difficulty in walking or climbing stairs; 1 indicates difficulty and 0 indicates no difficulty. |
| SEX | 0 for Female, 1 for Male. |
| AGE | Categorized into 13 levels based on 5-year intervals. |
| EDUCAG | Education level on a scale from 1 (lowest) to 6 (highest). |

## B. Machine Learning based prediction Results

The classification results for various machine learning models in predicting cardiovascular disease risk in diabetic patients show that XGBoost achieved the highest accuracy at 90.50%, with precision, recall, and F1-score all at 0.95, indicating excellent performance (see table IV). RF and AdaBoost also performed well, with accuracies of 90.47% and 90.35%, respectively. Other models, such as DT and KNN, also showed high accuracy, though slightly lower. Overall, XGBoost is the top-performing model in this analysis, demonstrating the best balance of accuracy and predictive metrics.

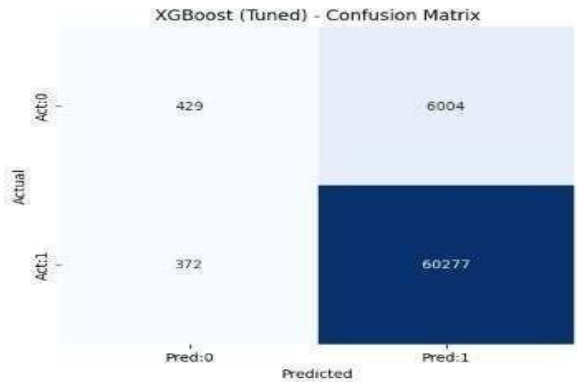

Fig. 2. Confusion matrix for XGBoost

TABLE IV
CLASSIFICATION RESULT FOR ML MODEL

| Model | Accuracy | Precision | Recall | F1-Score |
|---|---|---|---|---|
| DT | 0.9014 | 0.91 | 0.99 | 0.95 |
| RF | 0.9047 | 0.91 | 1.00 | 0.95 |
| AdaBoost | 0.9035 | 0.91 | 0.99 | 0.95 |
| KNN | 0.9009 | 0.91 | 0.99 | 0.95 |
| XGBoost | 0.9050 | 0.91 | 0.99 | 0.95 |

## C. Deep Learning based prediction Results

The classification results for DL models in predicting cardiovascular disease risk in diabetic patients reveal high performance across all models. The results indicate strong classification capabilities for each model. The results show that each model is very effective at classifying things. LSTM was the most accurate, with 90.50%. BiLSTM was next with 90.49%, while GRU was last with 90.47%. CNN also did well, getting 90.44% of the answers right and being the most accurate (0.9109). All three models, ANN, DNN, and RNN, had the same accuracy of 90.41% and perfect recall (1.0000). LSTM had the highest overall accuracy, and all of the models gave findings that were highly similar and reliable (see Table V).

TABLE V
CLASSIFICATION RESULT FOR DL MODEL

| Model | Accuracy | Precision | Recall | F1-Score |
|---|---|---|---|---|
| ANN | 0.9041 | 0.9041 | 1.0000 | 0.9496 |
| DNN | 0.9041 | 0.9041 | 1.0000 | 0.9496 |
| RNN | 0.9041 | 0.9041 | 1.0000 | 0.9496 |
| LSTM | 0.9050 | 0.9091 | 0.9944 | 0.9498 |
| GRU | 0.9047 | 0.9067 | 0.9973 | 0.9498 |
| BiLSTM | 0.9049 | 0.9084 | 0.9952 | 0.9498 |
| CNN | 0.9044 | 0.9109 | 0.9912 | 0.9494 |

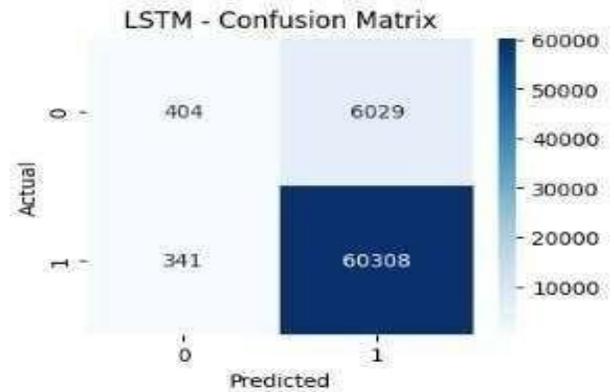

Fig. 3. Confusion matrix for LSTM

## D. Hybrid Model based prediction Results

The classification outcomes for hybrid models in predicting cardiovascular disease risk in diabetic patients demonstrate significant predictive validity. With an accuracy of 90.46%, a precision of 0.9095, and a recall of 0.9934, the CNN+LSTM model was the best. This conclusion means that it could identify positive cases and be reliable at the same time. The LSTM+GRU model had similar results, with an accuracy of 90.46%, but slightly less precision and a higher recall of 0.9981. CNN+BiLSTM and BiLSTM+GRU both had 90.44% and 90.41% accuracy, respectively. BiLSTM+GRU had a perfect recall of 1.00. In most cases, CNN+LSTM was the most accurate overall. The hybrid models, in their turn, all were high. This indicates that they achieved good precision, recall, and F1 scores in the process of generating predictions. (see Table VI).

TABLE VI
CLASSIFICATION RESULT FOR HYBRID MODEL

| Model | Accuracy | Precision | Recall | F1-Score |
|---|---|---|---|---|
| CNN+LSTM | 0.9046 | 0.9095 | 0.9934 | 0.9496 |
| CNN+LSTM | 0.9044 | 0.9070 | 0.9965 | 0.9496 |
| LSTM+GRU | 0.9046 | 0.9060 | 0.9981 | 0.9498 |
| BiLSTM+GRU | 0.9041 | 0.9041 | 1.0000 | 0.9496 |

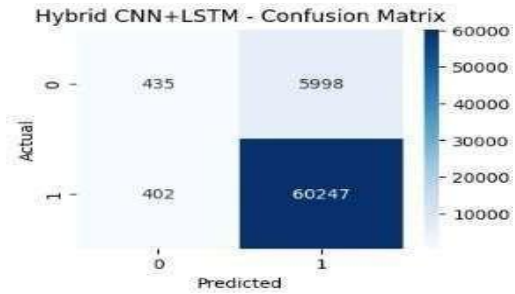

Fig. 4. Confusion matrix for CNN+LSTM

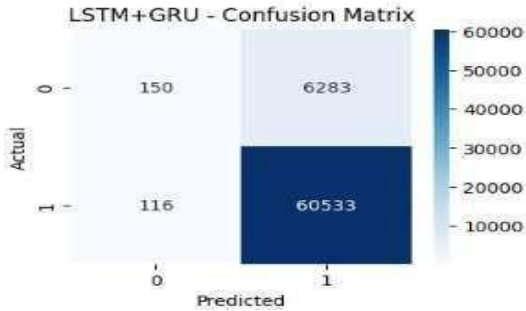

Fig. 5. Confusion matrix for LSTM +GRU

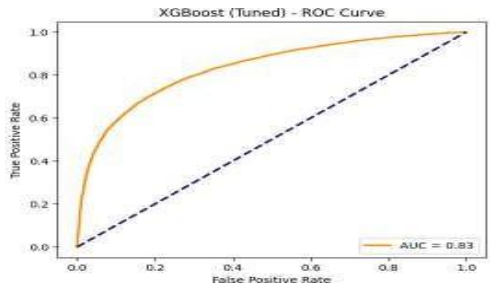

Fig.6: ROC curve of the XGBoost model

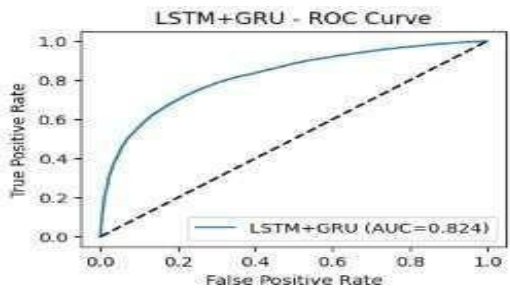

Fig.7: ROC curve of the LSTM+GRU

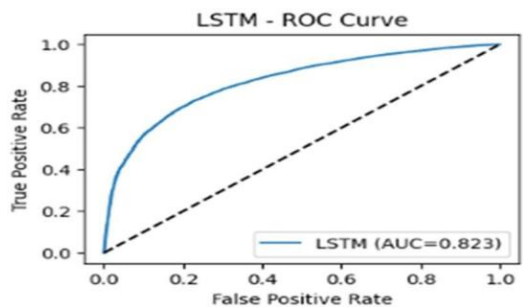

Fig.8: ROC curve of the LSTM model

## V. Conclusion and Future Work

This paper applies machine learning (ML) and deep learning (DL) methods to offer a holistic predictive model to evaluate the risk of cardiovascular disease among diabetics. We examined BRFSS data and used PCA to present additional information. We split it into 2 parts, 80% training and 20% testing. This strategy guaranteed the impartiality of the test and the effectiveness of the model. To maximize the accuracy and reliability of the predictions, this paper employed a number of different machine learning models, such as Decision Tree (DT), K-Nearest Neighbor (KNN), and XGBoost, along with artificial neural network (ANN), deep neural network (DNN), convolutional neural network (CNN), recurrent neural network (RNN), long short-term memory (LSTM), bidirectional long short-term memory (BiLSTM), gated recurrent unit (GRU), and hybrid architecture. The suggested procedure of evaluating the probability of heart problems in diabetics was always very precise, recalling F1 scores. The dataset may be refined in the future by adding more real-time patient data and better methods of ensemble and automated hyperparameter optimization. In addition, it can build a clinical support system that is understandable and user-friendly and hence increase its impact on healthcare.

## References

[1] H. Sang, H. Lee, M. Lee, J. Park, S. Kim, and H. G. Woo, *et al.*, "Prediction model for cardiovascular disease in patients with diabetes using machine learning derived and validated in two independent Korean cohorts," *Scientific Reports*, vol. 14, pp. 14966–14966, 2024.

[2] H. Calero-Diaz, D. Chushig-Muzo, H. Fabelo, I. Mora-Jiménez, C. Granja, and C. Soguero-Ruiz, "Data-driven cardiovascular risk prediction and prognosis factor identification in diabetic patients," *Proc. IEEE-EMBS Int. Conf. Biomed. Health Inform. (BHI)*, vol. 2022, pp. 1–4, 2022.

[3] M. Dorraki, Z. Liao, D. Abbott, P. J. Psaltis, E. Baker, N. Bidargaddi, A. van den Hengel, J. Narula, and J. W. Verjans, "Cardiovascular disease risk prediction via machine learning using mental health data," *European Heart Journal– Digital Health*, vol. 3, pp. 2784–2784, 2022.

[4] A. M. Johri, K. V. Singh, L. E. Mantella, L. Saba, A. Sharma, and J. R. Laird, *et al.*, "Deep learning artificial intelligence framework for multiclass coronary artery disease prediction using combination of conventional risk factors, carotid ultrasound, and intraplaque neovascularization," *Computers in Biology and Medicine*, vol. 150, pp. 106018–106018, 2022.

[5] W. Yang, Y. Guo, and Y. Liu, "Improvement of auxiliary diagnosis of diabetic cardiovascular disease based on data oversampling and deep learning," *Applied Sciences*, vol. 13, pp. 5449–5449, 2023.

[6] N. S. Punn and D. K. Dewangan, "Ensemble meta-learning using SVM for improving cardiovascular disease risk prediction," *medRxiv*, pp. 1–10, 2024.

[7] L. C. Jaffrin and J. Visumathi, "Impact of feature selection techniques on machine learning and deep learning techniques for cardiovascular disease prediction-an analysis," *Edelweiss Applied Science and Technology, Learning Gate*, vol. 8, no. 5, pp. 1454–1471, 2024.

[8] K. Vidhya and R. Shanmugalakshmi, "Deep learning based big medical data analytic model for diabetes complication prediction," *J. Ambient Intell. Human Comput.*, vol. 11, no. 4, pp. 5691–5702, 2020.

[9] A. R. Folsom, L. E. Chambless, B. B. Duncan, A. C. Gilbert, and J. S. Pankow, "Prediction of coronary heart disease in middle-aged adults with diabetes," *Diabetes Care*, vol. 26, no. 10, pp. 2777–2784, Oct. 2003.

[10] K. Zarkogianni, M. Athanasiou, A. C. Thanopoulou, and K. S. Nikita, "Comparison of machine learning approaches toward assessing the risk of developing cardiovascular disease as a long-term diabetes complication," *IEEE J. Biomed. Health Inform.*, vol. 22, no. 5, pp. 1637–1647, 2018.

[11] M. H. Alshayeji, S. Abed, and S. C. B. Sindhu, "Two-stage framework for diabetic retinopathy diagnosis and disease stage screening with ensemble learning," *Expert Syst. Appl.*, vol. 225, p. 120206, 2023.

[12] P. Thakur, V. Kansal, and V. Rishiwal, "Hybrid deep learning approach based on LSTM and CNN for malware detection," *Wireless Personal Communications*, vol. 136, no. 3, pp. 1879–1901, Jun. 2024.

[13] K. Papatheodorou, M. Banach, M. Edmonds, N. Papanas, and D. Papazoglou, "Complications of diabetes," *J. Diabetes Res.*, vol. 2015, pp. 1–5, 2015.

[14] A. S. Saldanha de Mattos Matheus, L. R. M. Tannus, R. A. Cobas, C. C. S. Palma, C. A. Negrato, and M. de B. Gomes, "Impact of diabetes on cardiovascular disease: an update," *Int. J. Hypertens.*, vol. 2013, pp. 1–15, 2013.

[15] L. Langouche and G. Van den Berghe, "Glucose metabolism and insulin therapy," *Crit. Care Clin.*, vol. 22, no. 1, pp. 119–129, 2006.

[16] J. A. Bluestone, K. Herold, and G. Eisenbarth, "Genetics, pathogenesis and clinical interventions in type 1 diabetes," *Nature*, vol. 464, no. 7293, pp. 1293–1300, 2010.

[17] S. E. Kahn, M. E. Cooper, and S. Del Prato, "Pathophysiology and treatment of type 2 diabetes: perspectives on the past, present, and future," *The Lancet*, vol. 383, no. 9922, pp. 1068–1083, 2014

[18] P. Z. Zimmet, D. J. Magliano, W. H. Herman, and J. E. Shaw, "Diabetes: a 21st century challenge," *Lancet Diabetes Endocrinol.*, vol. 2, no. 1, pp. 56–64, 2014.

[19] V. Lyssenko, A. Jonsson, P. Almgren, N. Pulizzi, B. Isomaa, and T. Tuomi, et al., "Clinical risk factors, DNA variants, and the development of type 2 diabetes," *New England Journal of Medicine*, vol. 359, no. 21, pp. 2220–2232, 2008.

[20] D. M. Lloyd-Jones, P. W. F. Wilson, M. G. Larson, A. Beiser, E. P. Leip, R. B. D'Agostino, and D. Levy, "Framingham risk score and prediction of lifetime risk for coronary heart disease," *Am. J. Cardiol.*, vol. 94, no. 1, pp. 20–24, Jul. 2004.